\documentclass[10pt,twocolumn,letterpaper]{article}

\usepackage[pagenumbers]{cvpr}

\usepackage{graphicx}
\usepackage{amsmath}
\usepackage{amssymb}
\usepackage{booktabs}
\usepackage{multirow}
\usepackage{makecell}
\usepackage{diagbox}
\usepackage{pifont}
\usepackage{xcolor}
\usepackage{colortbl}

\usepackage[pagebackref,breaklinks,colorlinks,citecolor=cvprblue]{hyperref}

\newlength\savewidth

\newcommand{\method}{ILLUME}

\usepackage[capitalize]{cleveref}
\crefname{section}{Sec.}{Secs.}
\Crefname{section}{Section}{Sections}
\Crefname{table}{Table}{Tables}
\crefname{table}{Tab.}{Tabs.}

\definecolor{cvprblue}{rgb}{0.21,0.49,0.74}

\title{\includegraphics[width=0.05\textwidth]{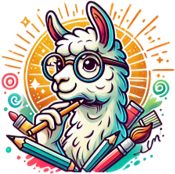} ILLUME: Illuminating Your LLMs to See, Draw, and Self-Enhance}

\author{Chunwei Wang$^{\ast}$, Guansong Lu$^{\ast}$, Junwei Yang$^{\ast}$, Runhui Huang, \\
Jianhua Han, Lu Hou, Wei Zhang, Hang Xu$^{\dag}$ \\
Huawei Noah’s Ark Lab\\
}

\begin{document}
\maketitle
\footnotetext{$^{\ast}$Equal contribution, $^{\dag}$Corresponding author} 
\begin{abstract}
\looseness=-1
In this paper, we introduce \method, a unified multimodal large language model (MLLM) that seamlessly integrates multimodal understanding and generation capabilities within a single large language model through a unified next-token prediction formulation.
To address the large dataset size typically required for image-text alignment, we propose to enhance data efficiency through the design of a vision tokenizer that incorporates semantic information and a progressive multi-stage training procedure. This approach reduces the dataset size to just 15M for pretraining -- over four times fewer than what is typically needed -- while achieving competitive or even superior performance with existing unified MLLMs, such as Janus. Additionally, to promote synergistic enhancement between understanding and generation capabilities, which is under-explored in previous works, we introduce a novel self-enhancing multimodal alignment scheme. This scheme supervises the MLLM to self-assess the consistency between text descriptions and self-generated images, facilitating the model to interpret images more accurately and avoid unrealistic and incorrect predictions caused by misalignment in image generation. 
Based on extensive experiments, our proposed \method~stands out and competes with state-of-the-art unified MLLMs and specialized models across various benchmarks for multimodal understanding, generation, and editing.
\end{abstract}
\section{Introduction}
Recent research efforts~\cite{liu2024visual,liu2024improved,bai2023qwen,chen2024internvl} have equipped Large Language Models (LLMs) with the capability to ``see" images by utilizing vision adapters to map features from CLIP-like encoders into LLM's input spaces. Works like the LLaVA series~\cite{liu2024improved,liu2024llava} have demonstrated exceptional results on visual comprehension tasks.
Meanwhile, the field of text-to-image generation has achieved remarkable progresses in developing both diffusion-based~\cite{peebles2023scalable,podell2023sdxl} and more recent autoregressive models~\cite{sun2024autoregressive,he2024mars}. 
These technological strides are propelling the community towards the creation of versatile Multimodal Large Language Models (MLLMs) that can seamlessly integrate visual understanding and generation capabilities. 
This integration not only extends LLMs across a wide range of multimodal applications, but also unlock new possibilities for improving the synergy between vision and language tasks.

\begin{figure}[t!]
    \begin{center}
        \includegraphics[width=0.85\linewidth]{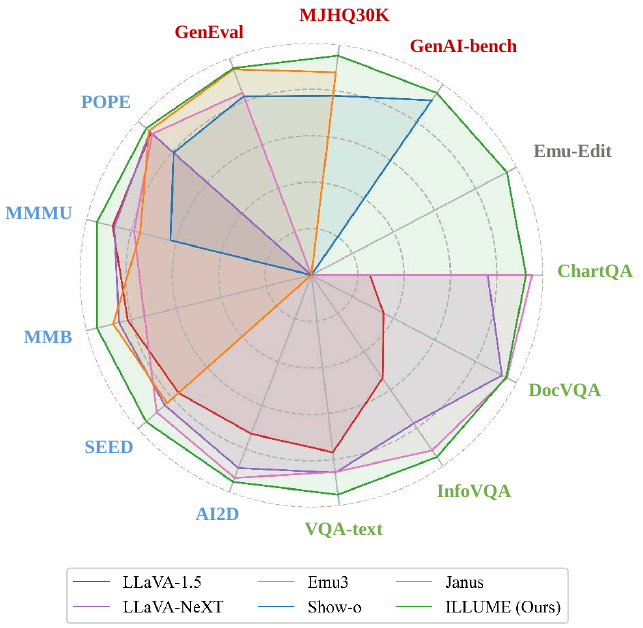}
    \end{center}
    \vspace{-4mm}
    \caption{Performance on various visual understanding (blue for General and green for Document-oriented), generation (red), and editing (gray) benchmarks. \method~achieves competitive results with state-of-the-art works.}
    \label{fig-benchmark}
    \vspace{-4mm}
\end{figure}

\begin{figure*}[htbp]
    \begin{center}
        \includegraphics[width=1.0\linewidth]{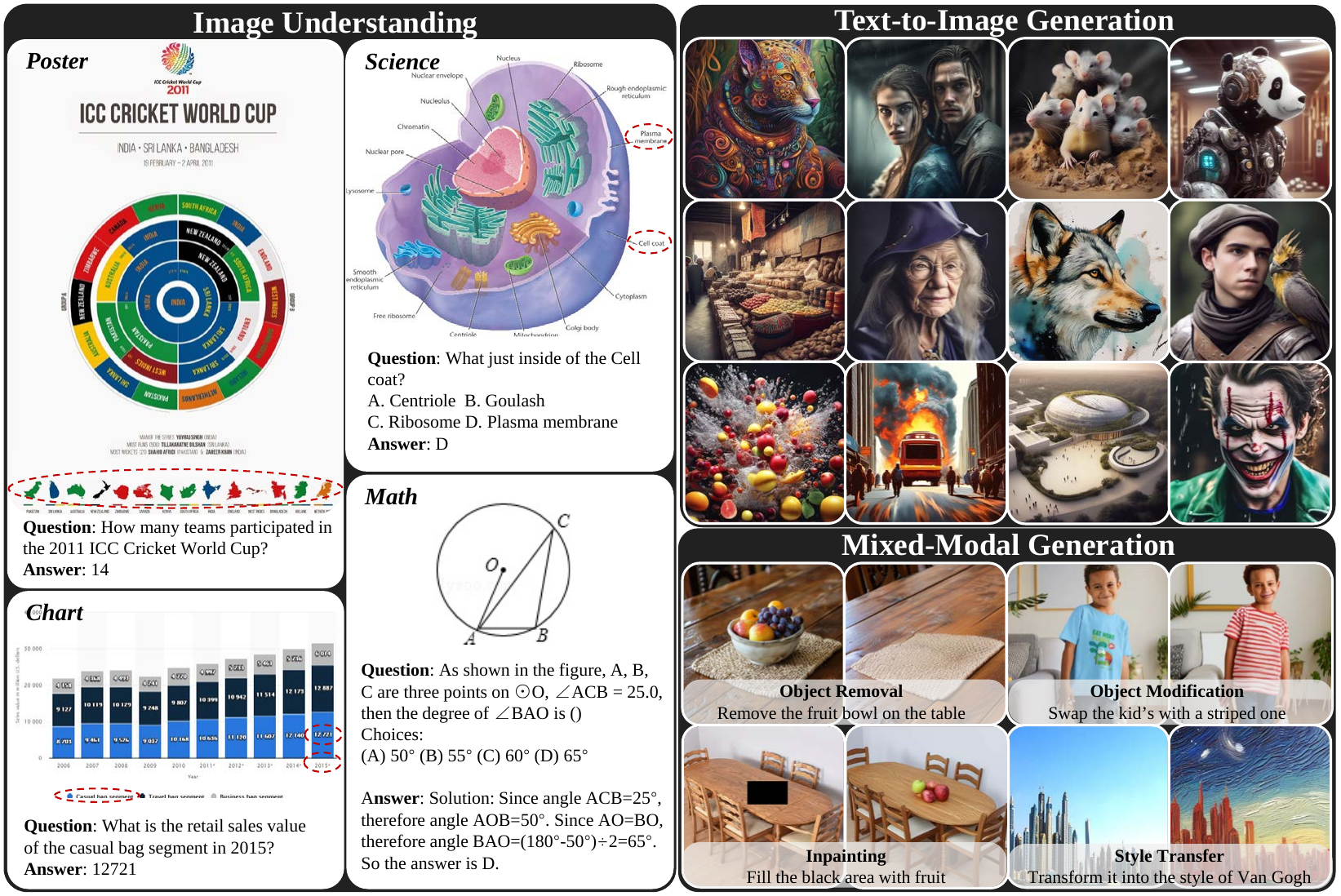}
    \end{center}
    \vspace{-2mm}
    \caption{\method~can handle various multimodal tasks, including understanding for images and charts; text-to-image generation; and mixed-modal generation task such as object modification and style transfer. 
    }
    \label{fig-vis}
    \vspace{-4mm}
\end{figure*}

Previous works have explored several methods that empower MLLMs to ``draw'' images. 
In \cite{wu2023visual,li2024mini}, image generation models are deployed as tools, with text commands as input to create images. This decoupled architecture inherently limits the models' potential.
In contrast, Emu~\cite{sun2023generative}, Emu2~\cite{sun2024generative}, and X-VILA~\cite{ye2024x} introduce a unified autoregressive model that alternatively predicts the next element by regressing visual embeddings or classifying text tokens. 
While these models have shown promising results, joint training of LLMs and diffusion models requires substantial engineering costs and well-designed training strategy to ensure stability.
To avoid these complexities, recent innovations like Chameleon ~\cite{team2024chameleon}, AnyGPT~\cite{zhan2024anygpt} and Emu3~\cite{wang2024emu3} employ a Vector-Quantized (VQ) tokenizer to transform images into discrete tokens and extends LLMs with an additional vision vocabulary. In this paradigm, the MLLM is optimized via a unified next-token prediction formulation. Along with the consistency of the discrete design with text,  these approaches open up substantial potential for multimodal models.

However, we observe that extending the vision vocabulary in an LLM necessitates extensive data for image-text alignment in existing methods, as indicated in Table~\ref{tab:statistic}. This observation prompts us to ask: \textbf{Can we develop a unified MLLM more efficiently?} In response, we propose \method, a unified MLLM that requires only 15M data for image-text alignment during MLLM pretraining -- four times fewer than Janus~\cite{wu2024janus} -- yet delivers competitive performance compared to state-of-the-art models.
This increased efficiency is primarily attributed to two designs. First, we employ a semantic vision tokenizer tailored for MLLMs. Unlike traditional VQ tokenizers that rely on image reconstruction loss for training (e.g., VQGAN~\cite{esser2021taming}), our approach quantizes images into discrete tokens within a semantic feature space. This method significantly accelerates the image-text alignment process in MLLMs. 
Moreover, \method~ is implemented with a three-stage training procedure. It innovatively introduces an image reconstruction task to facilitate rapid initialization of the newly integrated weights in LLMs due to the extension of vision vocabulary, promoting the model to learn pixel dependencies for image generation.
With a diverse range of vision-language data types utilized during training, \method~is ultimately capable of handling various multimodal tasks, as illustrated in Figure~\ref{fig-vis}.
\begin{table}[t]
\center
\setlength{\tabcolsep}{3.6pt}
\Large
\resizebox{0.48\textwidth}{!}{
\begin{tabular}{lccc}
\toprule
Method & LLM & Num. of image-text pairs &  Num. of interleaved data \\
\midrule
Chameleon~\cite{team2024chameleon} & 7B from scratch &  1.4B & 400B tokens\\
LWM~\cite{liu2024world} & LLaMA-2-7B & 1B & - \\
Unified IO 2~\cite{lu2024unified} & 6.8B from scratch & 970M & 157M \\
SEED-LLaMA~\cite{ge2023making} & Vicuna-7B & 600M &  150M\\
AnyGPT~\cite{zhan2024anygpt} & LLaMA-2 7B & 300M & 7.3M \\
Janus~\cite{wu2024janus} & DeepSeek-LLM-1.3B & 65M & - \\
\midrule
ILLUME (Ours) & Vicuna-7B & 15 M & - \\
\bottomrule
\end{tabular}
}
\vspace{-2mm}
\caption{Statistics on the data volumes required for image-text alignment in previous next-token prediction-based works. Notably, \method~ utilizes only 15M image-text pairs, which is 4 times fewer than Janus, yet achieves superior performance.
}
\vspace{-4mm}
\label{tab:statistic}
\end{table}

\begin{figure*}[t]
    \begin{center}
        \includegraphics[width=1.0\linewidth]{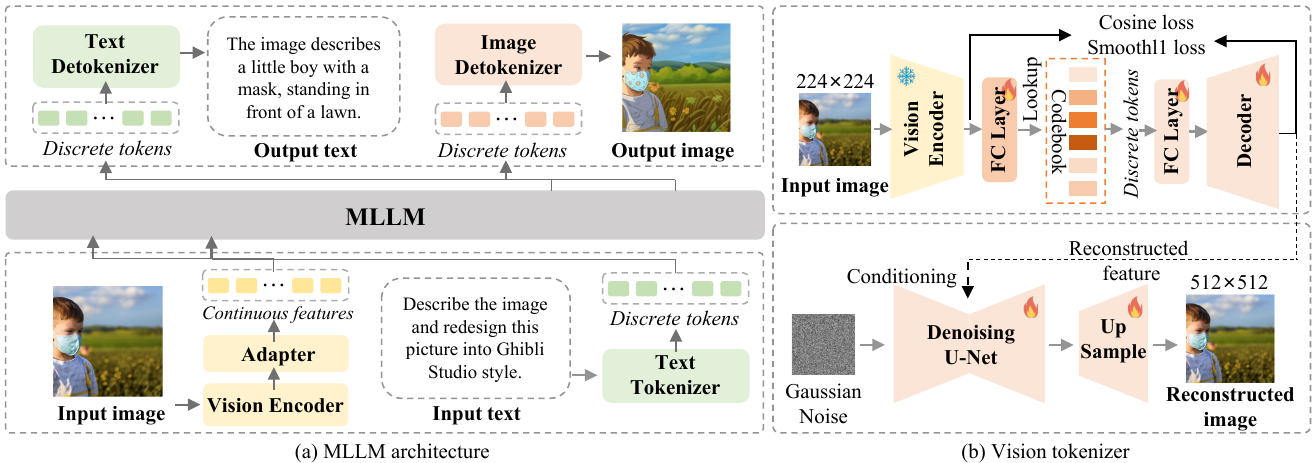}
    \end{center}
    \vspace{-5mm}
    \caption{Overall architecture of \method. (a) We enhance LLMs with the capability to ``see'' images by employing a vision adapter that maps features from a vision encoder into LLM’s input spaces. To expand the model’s abilities to generate images, the LLM is extended with an additional vision vocabulary to produce discrete vision tokens. (b) In the vision tokenizer, we utilize a pretrained vision encoder to extract semantic features and supervise quantization process through feature reconstruction loss. The reconstructed features are then processed by a Stable Diffusion model to recover the original images.}
    \label{fig-framework}
    \vspace{-4mm}
\end{figure*}

When we obtained our unified model, a new question arises: \textbf{Can the discriminative and generative capabilities of an MLLM enhance each other?}  Previous research on LLMs~\cite{chen2023gaining} suggests that self-generated contents can serve as valuable feedback for the model to self-improve.
The ability of an MLLM to generate images allows it to pinpoint and address its
weaknesses by learning from its own imperfect outputs, thereby enhancing its ability to interpret images more accurately.
Furthermore, the MLLM can utilize its discriminative skills to evaluate whether self-generated images align with user instructions, guiding it to avoid potential mistakes in generating images.
To harness this potential, we propose a novel self-enhancing multimodal alignment scheme that teaches MLLMs to assess the consistency between self-generated images and text descriptions, as well as to understand the underlying reasons for any discrepancies.
With this alignment scheme, we observe improvements in both discriminative and generative capabilities within a unified infrastructure. We evaluate our model on popular visual understanding, generation and editing benchmarks, where \method~achieves competitive results with existing unified MLLMs and specialized models.

In brief, our contributions are summarized as follows.
\begin{itemize}
    \item We introduce \method, a unified MLLM that seamlessly integrates visual understanding and generation within a single LLM, which is efficiently trained with the aid of a semantic vision tokenizer and a three-stage procedure. 
    \item To promote synergistic enhancement between understanding and generation capabilities, we introduce a novel self-enhancing multimodal alignment scheme that trains MLLMs to self-assess the consistency between text descriptions and self-generated images. 
    \item \method~excels among existing unified MLLMs and exhibits  competitive performance compared to specialized models across a diverse range of benchmarks in multimodal understanding, generation, and editing.
\end{itemize}
 
\section{Related Work}
\paragraph{Multimodal Understanding.} 
The significant advancements in LLMs have spurred the development of Large Vision-Language Models (LVLMs). To bridge the gap between vision and text modalities, early approaches such as LLaVA~\cite{liu2024visual} and MiniGPT-4~\cite{zhu2023minigpt} utilize vision adapters to align vision features from vision encoders to the input space of LLMs. 
Further improvements have been observed in models such as the LLaVA series~\cite{liu2024improved,liu2024llava}, Qwen series~\cite{bai2023qwen,wang2024qwen2}, and InternVL series~\cite{chen2024internvl,chen2024far}, which are achieved through the use of higher-quality datasets, increased input image resolution, and enhanced training strategies. Despite their strong understanding capabilities, these models support only visual perception and comprehension tasks. 

\begin{figure*}[t]
    \begin{center}
        \includegraphics[width=1.0\linewidth]{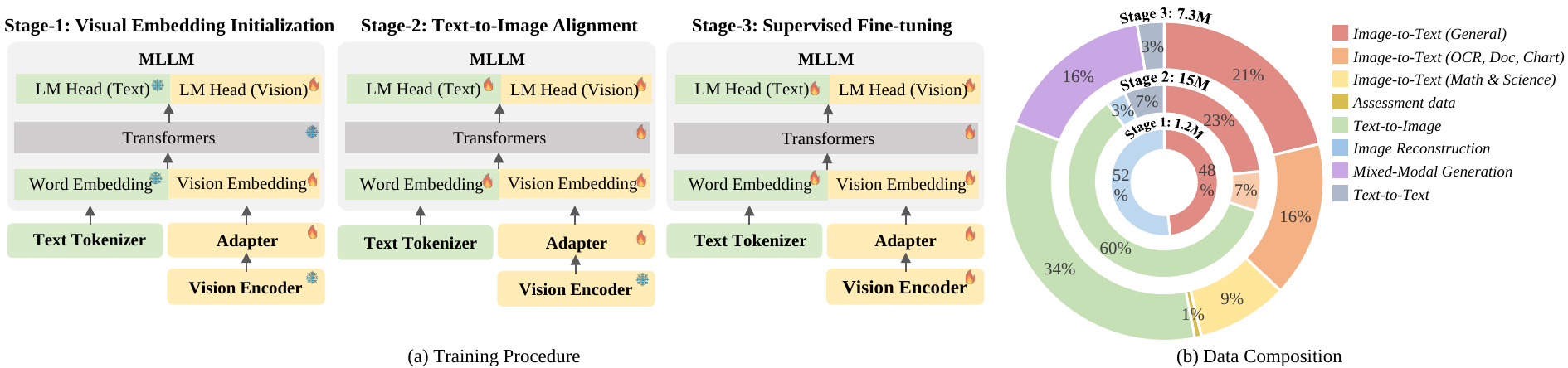}
    \end{center}
    \vspace{-5mm}
    \caption{Overview of the three-stage training procedure and its corresponding data composition of different stages in MLLM training. 
    }
    \label{fig-training_procedure}
    \vspace{-4mm}
\end{figure*}

\vspace{-2mm}
\paragraph{Visual Generation.}
Diffusion-based methods~\cite{rombach2022high,peebles2023scalable,podell2023sdxl,ruiz2023dreambooth} have shown exceptional capabilities and have become dominant in the image generation domain in recent years. These models operate by predicting Gaussian noise in a forward process, and then generating high-quality images through an inverse denoising process. 
Another line of research~\cite{ramesh2021zero,ding2021cogview,yu2022scaling} converts images into discrete tokens using VQGAN-like vision tokenizers~\cite{esser2021taming,lee2022autoregressive}, and generates images by predicting the next token in a sequence autoregressively. In this paper, we extend the capabilities of MLLMs to image generation tasks using a unified autoregressive form, and further adopt diffusion model to reconstruct high-quality images from the predicted tokens.

\vspace{-2mm}
\paragraph{Unified Visual Understanding and Generation.}
An increasing number of studies are making efforts to unify visual understanding and generation tasks with LLMs. Pioneering works such as Emu~\cite{sun2023generative}, Emu2~\cite{sun2024generative}, and X-VILA~\cite{ye2024x} develop a unified autoregressive model that predicts the next multimodal element, either by regressing visual embeddings or classifying text tokens. Yet, the non-unified design of optimization goals for different modalities may limit the feature integration across modalities, while the joint training with an extra diffusion decoder further complicates the infrastructure design with overall lower efficiency.
On the other hand, methods like LWM~\cite{liu2024world}, AnyGPT~\cite{zhan2024anygpt}, Show-o~\cite{xie2024show}, and VILA-U~\cite{wu2024vila} utilize a VQ tokenizer to transform images into vision tokens, enabling LLMs to be optimized via a unified loss to predict the next token in both text and vision contexts. In this work, we explore methods for both the efficient text-image alignment in MLLMs and the synergy of discriminative and generation capabilities, which are under exploration in previous works.

\section{\method}
This section presents our proposed framework \method, a unified model for visual understanding and generation. More specifically, details of the design of vision tokenizer, MLLM, and training procedures are discussed. 

\subsection{Vision Tokenizer}
To process input images in LLMs, previous VLMs such as LLaVA \cite{liu2024visual}, have demonstrated efficient text-image alignment by utilizing a vision adapter to map semantic features from vision encoder to text space, utilizing only a dataset of 558K samples for pretraining.
However, in the domain of image generation, most existing autoregressive-based unified models~\cite{team2024chameleon,zhan2024anygpt} are struggling with extensive training data required for LLM pretraining, as detailed in Table~\ref{tab:statistic}. We hypothesize that this issue stems from the inadequate semantic information provided by current vision tokenizers, such as VQGAN~\cite{esser2021taming}, which are not optimally suited for LLMs. These tokenizers are trained primarily on image reconstruction loss, with visual representation focusing on low-level textures for quantization, which in turn hampers text-image alignment in MLLMs. To this end, we resort to quantizing images into discrete tokens within a semantic feature space. Specifically, as illustrated in Figure~\ref{fig-framework}(b), we utilize UNIT~\cite{zhu2024unit}, a pretrained vision encoder, to extract semantic features and supervise the quantization process along with codebook learning through feature reconstruction loss. This approach significantly accelerates the image-text alignment process in comparison with those tokenizers with image reconstruction loss, as demonstrated by the observations in Figure~\ref{fig-preliminary_tokenizer}.

Moreover, since quantization occurs within a semantic feature space, we further utilize the Stable Diffusion (SD) model~\cite{podell2023sdxl} to reconstruct these semantic features back into images with a high compression ratio of $32\times$. The robust SD model effectively compensates for the low-level details that are not preserved during the quantization process. This allows for the generation of higher-resolution images from a fixed number of discrete tokens output by the MLLMs. 

\begin{figure*}[t]
    \begin{center}
        \includegraphics[width=1.0\linewidth]{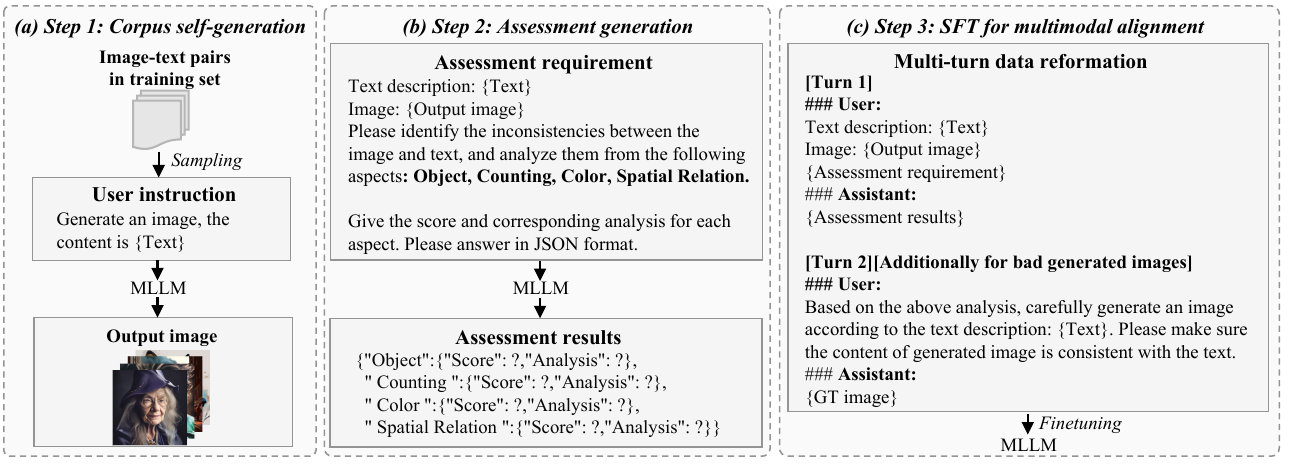}
    \end{center}
    \vspace{-4mm}
    \caption{Procedure of self-enhancing multimodal alignment scheme, which contains three steps: corpus self-generation, assessment generation and SFT for multimodal alignment. This scheme supervises the MLLM to self-assess the consistency between text descriptions and self-generated images, enabling the model to more accurately interpret images and avoid potential mistakes in image generation.}
    \label{fig-self_enhanced}
    \vspace{-4mm}
\end{figure*}

\subsection{MLLM}
\paragraph{Architecture.} As shown in Figure~\ref{fig-framework}, \method~inherits the architecture of existing Visual Language Models (VLMs)~\cite{liu2024improved,liu2024llava} by extending LLMs with an additional vision vocabulary to generate discrete vision tokens. For visual understanding, we utilize the UNIT encoder~\cite{zhu2024unit}, which is utilized in our proposed vision tokenizer, to extract semantic features from input images. These features are further aligned to the input space of the LLM via a vision adaptor. This design mitigates the information loss caused by vector quantization, which is vital for tackling fine-grained multimodal understanding tasks.
For visual generation, we use our vision tokenizer to convert images into discrete indices, and supervise the token prediction at each location for both modalities with a shared prediction head in LLMs. With this architecture, \method~adopts the general Language Modeling (LM) objective to directly maximize the likelihood of each multi-modal sequence in an auto-regressive manner: 
\begin{equation}
    \mathcal{L} = -\sum_{i=1}\log P_{\theta}(y_i|y_{\leq i}),
\end{equation}
where $y_i$ represents the text or visual token, and $\theta$ denotes the parameters of the LLM.
Notably, given our model's capability to handle images in both the input and output, our proposed framework is compatible with interleaved image-text data to support any-to-any multimodal tasks.

\paragraph{Training Procedure.}
The training procedure and data composition of MLLM is illustrated in Figure~\ref{fig-training_procedure}. 
The training procedure consists of three stages as below.
\begin{itemize}
    \item \textit{Stage-1: Visual Embedding Initialization.} The primary goal of this stage is to initialize a good visual representation for subsequent training steps. The vision adaptor is trained by leveraging image-to-text pairs from LLaVA-Pretrain~\cite{liu2024visual} to transform the visual features from vision encoder into LLMs' text space. 
    Meanwhile, this stage also involves the learning of new learnable embeddings, where only the vision embedding and the vision part of final classifier head of the LLM are updated.    
    We introduce image reconstruction task, i.e., supervising the LLM to generate the original images, to facilitate rapid initialization of the introduced integrated weights in LLMs.
    \item \textit{Stage-2: Unified Image-Text Alignment.}
    This stage focus on image-text alignment to learn both the understanding and generation tasks on multimodal data. We unfreeze the LLM and vision adaptor, utilizing 15M training data for training, including text data, image caption data for both natural images and documents, image data for reconstruction, and text-to-image generation data.
    \item \textit{Stage-3: Supervised Fine-tuning.} 
    After pretraining, we train the whole model with task-specific data to handle various multimodal understanding and generation tasks.
    To receive high-resolution images for fine-grained multimodal understanding like OCR and document-oriented tasks, we employ image patchfy strategy following LLaVA-NeXT~\cite{liu2024llava}. 
    This stage utilizes instruction tuning data following \cite{chen2024emova} for visual understanding, high-quality image-text pairs for text-to-image generation and various mixed-modal generation data. 
\end{itemize}

\paragraph{Inference.}
During inference, our model adopts the next-token prediction approach. For visual understanding, we follow the standard practice of sampling tokens sequentially from the predicted distribution. For image generation, we utilize classifier-free guidance (CFG) as used in prior works~\cite{liu2024lumina,xie2024show}.

\begin{table}[t]
\center
\setlength{\tabcolsep}{3.6pt}
\Large
\resizebox{0.48\textwidth}{!}{
\begin{tabular}{l|cc|ccccccc}
\toprule
Tasks  & GenAI-Bench & GenEval & POPE & MME-P & MMBench & SEED & MMVet \\
\midrule
Gen. only & 0.63 & 0.58 & - & - & - & - \\ 
Und. only & - & - & 84.6 & 1339.0 & 60.9 & 64.0 & 28.0 \\ 
Gen. and Und. & 0.63 & 0.56 & 86.4 & 1358.6 & 61.6 & 65.0 & 27.4\\ 
\bottomrule
\end{tabular}
}
\vspace{-2mm}
\caption{Comparison between the specialist model and unified model. Joint training presents no significant negative impact on the two tasks, but it also does not obviously promote each other.
}
\vspace{-4mm}
\label{tab:split_or_joint}
\end{table}

\begin{table*}[t]
\center
\setlength{\tabcolsep}{3.6pt}
\Large
\resizebox{\textwidth}{!}{
\begin{tabular}{ll|ccccccc|ccccc}
\toprule
 &   & \multicolumn{7}{c|}{\textbf{General}} & \multicolumn{5}{c}{\textbf{Doc}} \\
\multirow{-2}[0]{*}{Method} & \multirow{-2}[0]{*}{LLM.} & POPE & MMBench & SEED & MME-P & MM-Vet & MMMU & \multicolumn{1}{c|}{AI2D} & VQA-text & ChartQA & DocVQA & InfoVQA & OCRBench \\
\midrule
\multicolumn{14}{c}{\textbf{\it Understanding Only}} \\
\midrule
InstructBLIP~\cite{instructblip} & Vicuna-7B & - & 36.0 & 53.4 & - & 26.2  & 30.6  & 33.8 & 50.1 & 12.5  & 13.9  & - & 276 \\
Qwen-VL-Chat~\cite{bai2023qwen} & Qwen-7B & - & 60.6 & 58.2 & 1487.5 & -  & 35.9 & 45.9 & 61.5 &  66.3  & 62.6 & - & 488 \\
LLaVA-1.5~\cite{liu2024improved} & Vicuna-7B & 85.9 & 64.3 & 58.6 & \underline{1510.7}  & 31.1 & 35.4 & 54.8 & 58.2 & 18.2  & 28.1  & 25.8 & 318 \\
ShareGPT4V~\cite{chen2023sharegpt4v} & Vicuna-7B & - & 68.8 & \underline{69.7} & \textbf{1567.4} & \underline{37.6}  & \underline{37.2}  & 58  & 60.4 & 21.3  & -  & - & 371 \\
LLaVA-NeXT~\cite{liu2024llava} & Vicuna-7B & 86.5 & 67.4 & 64.7 & - & \textbf{43.9} & 35.1 & 66.6 & \underline{64.9} & 54.8  & 74.4  & 37.1 & 532 \\
Emu3-Chat~\cite{wang2024emu3} & 8B from scratch & 85.2 & 58.5 & 68.2 & - & 37.2  & 31.6  & \underline{70.0} & 64.7 & \textbf{68.6} & \textbf{76.3}  & \underline{43.8} & \textbf{687} \\
\midrule
\multicolumn{14}{c}{\textbf{\it Unify Understanding and Generation}} \\
\midrule
Unified-IO 2~\cite{lu2024unified} & 6.8B from scratch & \underline{87.7} & -  & 61.8 & - & - & - & - & - & -  & - & - & - \\
Chameleon~\cite{team2024chameleon} & 7B from scratch & - & - & - & - & 8.3 & 22.4 & -  & -  & - & - & - & - \\
LWM~\cite{liu2024world} & LLaMA-2-7B & 75.2 & - & - & - & 9.6 & - & - & 18.8 & -  & - & - & -   \\
Show-o~\cite{xie2024show} & Phi-1.5B & 73.8 & - & - & 948.4 & - & 25.1 & - & - & -  & - & - & -  \\
VILA-U (256)~\cite{wu2024vila} & LLaMA-2-7B & 83.9 & - & 56.3 & 1336.2 & 27.7 & - & -  & 48.3 & -  & - & - & -  \\
VILA-U (384)~\cite{wu2024vila} & LLaMA-2-7B & 85.8 & - & 59 & 1401.8 & 33.5 & - & -  & 60.8 & -  & - & - & -  \\
Janus~\cite{wu2024janus} & DeepSeek-LLM-1.3B & 87.0 & \underline{69.4} & 63.7 & 1338.0 & 34.3 & 30.5 & -  & - & - & - & - & -  \\
\midrule
\rowcolor[gray]{0.9}ILLUME (Ours) & Vicuna-7B & \textbf{88.5} & \textbf{75.1} & \textbf{72.9}& 1445.3 & 37.0 & \textbf{38.2} & \textbf{71.4} & \textbf{72.1} & \underline{66.7} & \underline{76.0} & \textbf{45.5} & \underline{669} \\
\bottomrule
\end{tabular}
}
\vspace{-2mm}
\caption{\textbf{Quantitative results on visual understanding benchmarks.} Our performance is close to and even outperforms both understanding only and unified models. The performance with top-1 and top-2 value are denoted in bold and underline respectively.}
\vspace{-2mm}
\label{tab:understanding_results}
\end{table*}

\section{Self-Enhancing Multimodal Alignment}
The primary goal of our community in developing a unified MLLM is twofold: first, it can be easily extended to various vision-language tasks; second, the complete unification of representation spaces facilitates a more efficient learning process through better multimodal interaction and alignment. Therefore, once we build our ~\method, our priority was to investigate whether jointly improving these capabilities could benefit from the commonalities across the knowledge required for each one. However, according to our experimental results as shown in Table~\ref{tab:split_or_joint}, while there was no significant negative impact from joint training, the anticipated mutual enhancement between understanding and generation was not presented on existing benchmarks. This outcome underscores that while these capabilities can coexist without detrimental effects, their synergistic potential may require further exploration and more refined approaches.

In this work, we introduce a novel self-enhancing multimodal alignment scheme, as depicted in Figure~\ref{fig-self_enhanced}, which employs a self-assessment process as a bridge to synergistically enhance the discriminative and generative capabilities. We posit that if an MLLM can learn to assess the quality of its self-generated images during training, it can benefit in two aspects:
\begin{itemize}
    \item \textbf{Generation Aids Discrimination: }By analyzing self-generated negative samples, the MLLM learns to identify and understand its failures, thereby enhancing its ability to interpret images more accurately. This introspective process allows the model to pinpoint and address its weaknesses through self-assessment, leading to improved understanding and fewer misinterpretations.
    \item \textbf{Discrimination Aids Generation: } The MLLM could utilize its discriminative skills to assess whether its self-generated images align with texts, making necessary adjustments based on this analysis. This capability ensures that during inference, the model is more cautious and precise, avoiding potential mistakes in generating images.
\end{itemize} 

Inspired by the above motivation, we design a self-enhancing multimodal alignment scheme, which comprises three steps:
\begin{itemize}
    \item \textit{Step 1: Corpus self-generation. } The model self-generates images from a subset of text-to-image data within the training set.
    \item \textit{Step 2: Assessment generation. } We assess the inconsistencies between image and text against specific criteria such as object accuracy, count, color, and spatial relations. During generation, not only the assessment score (i.e., good or bad), but also the corresponding analysis are included. To obtain high-quality data, we resort to GPT4-o for assessment data generation with the template in Figure~\ref{fig-self_enhanced}(b).
    \item \textit{Step 3: SFT for multimodal alignment. } we reformat the assessment data as depicted in Figure~\ref{fig-self_enhanced}(c). Specifically, for instances identified as ``good generation cases'', we structure the data to only undergo the first round for assessment. As for ``bad generation cases'',  we reconstruct the data to two rounds of conversations, where the first round for assessment and the second round for refinement. In total, 50K assessment data are created with this scheme and we incorporate it into the Stage-3 of our training process.
\end{itemize} 

\section{Experiments}
We evaluate the proposed \method~on various multimodal understanding and generation benchmarks, and conduct  ablation studies to verify our design choices.

\subsection{Implementation Details}
In our experiments, we utilize Vicuna-7B as the base language model. For the vision encoder used in understanding tasks, we select UNIT~\cite{zhu2024unit}. The input image resolution is set as $224$ in Stage-1 and Stage-2, with $256$ token per image for LLMs. In Stage-3, we employ the image patchfy strategy following \cite{liu2024llava} to support high resolution images as input for fine-grained understanding, with a maximum slice number of $9$ and the base resolution of $448$. Each sliced image is downsampled to $256$ tokens. For image generation, the vision tokenizer has a codebook of size of $16384$, where the generated image has the resolution of $512\times512$ with $256$ discrete tokens. The training hyperparameters are illustrated in Table~\ref{tab:hyperparams}. The whole training process took $3$ days on a cluster of 32 nodes, each equipped with 8 Ascend NPUs. 

\begin{table*}[t]
\center
\setlength{\tabcolsep}{3.6pt}
\Large
\resizebox{\textwidth}{!}{
\begin{tabular}{lcc|c|cc|ccccccc}
\toprule
 &   &  & \multicolumn{1}{c|}{\it MJHQ30k} & \multicolumn{2}{c|}{\it GenAI-bench} & \multicolumn{7}{c}{\it GenEval} \\
\multirow{-2}[0]{*}{Method} & \multirow{-2}[0]{*}{Params.} & \multirow{-2}[0]{*}{Type} & \multicolumn{1}{c|}{FID} & Basic & \multicolumn{1}{c|}{Advanced} & Overall & Single Obj & Two Obj. & Counting & Colors & Position & Color Attri. \\
\midrule
\multicolumn{13}{c}{\textbf{\it Generation Only}} \\
\midrule
SDv1.5~\cite{rombach2022high} & 0.9B & Diffusion & - & - & - & 0.43 & 0.97 & 0.38 & 0.35 & 0.76  & 0.04 & 0.06 \\
PixArt-$\alpha$~\cite{chen2023pixart} & 0.6B & Diffusion & 6.14 & - & - & 0.48 & 0.98 & 0.50 & 0.44 & 0.80  & 0.08 & 0.07  \\
SDXL~\cite{podell2023sdxl} & 2.6B & Diffusion & 9.55 & 0.83 & 0.63 & 0.55 & 0.98 & 0.74 & 0.39 & 0.85  & 0.15 & 0.23  \\
Emu3-Gen~\cite{wang2024emu3} & 8B & Autoregressive & - & - & - & 0.54 & 0.98 & 0.71 & 0.34 & 0.81  & 0.17 & 0.21  \\
\midrule
\multicolumn{13}{c}{\textbf{\it Unify Understanding and Generation}} \\
\midrule
Chameleon~\cite{team2024chameleon} & 7B & Autoregressive & - & - & - & 0.39 & - & - & - & -  & -  & -  \\
LWM~\cite{liu2024world} & 7B & Autoregressive & 17.77 & 0.63 & 0.53 & 0.47 & 0.93 & 0.41 & \underline{0.46} & 0.79  & 0.09  & 0.15  \\
Show-o~\cite{xie2024show} &  1.5B & Autoregressive & 15.18 & 0.70 & 0.60 & 0.53 & 0.95 & 0.52 & \textbf{0.49} & \underline{0.82}  & 0.11  & \underline{0.28}  \\
VILA-U(256)~\cite{wu2024vila} &  7B & Autoregressive & 12.81 & \textbf{0.76} & \textbf{0.64} & - & - & - & - & -  & -  & -  \\
VILA-U(384)~\cite{wu2024vila} & 7B & Autoregressive & \textbf{7.69} & 0.73 & \underline{0.61} & - & - & - & - & -  & -  & -  \\
Janus~\cite{wu2024janus} & 1.3B & Autoregressive & 10.10 & - & - & \textbf{0.61} & \underline{0.97} & \underline{0.68} & 0.30 & \textbf{0.84}  & \textbf{0.46}  & \textbf{0.42} \\
\midrule
\rowcolor[gray]{0.9}ILLUME (Ours) & 7B & Autoregressive & \underline{7.76} & \underline{0.75} & 0.60 & \textbf{0.61} & \textbf{0.99} & \textbf{0.86} & 0.45 & 0.71 & \underline{0.39} & \underline{0.28} \\
\bottomrule
\end{tabular}
}
\vspace{-2mm}
\caption{\textbf{Quantitative results on text-to-image generation benchmarks.} \method~achieves comparable results with specialist models and unified MLLMs. The performance with top-1 and top-2 value are denoted in bold and underline respectively.
}
\vspace{-4mm}
\label{tab:generation_results}
\end{table*}
\begin{table}[t]
\center
\setlength{\tabcolsep}{3.6pt}
\Large
\resizebox{0.48\textwidth}{!}{
\begin{tabular}{cccc}
\toprule
Setting & Stage-1 & Stage-2 &  Stage-3 \\
\midrule
 & Vision adapter $1.0\times 10^{-3}$ & Vision adapter $5.0\times 10^{-5}$ & Vision encoder $2.0\times 10^{-6}$ \\
\multirow{-2}[0]{*}{LR.} &  Vision Embed. \& Head $2.0\times 10^{-4}$  & LLM $5.0\times 10^{-5}$ & LLM \& Vision adapter $2.0\times 10^{-5}$ \\

Batch size & 256 & 1024 & 1024 \\
Training Step & 5000 & 15000 & 8000 \\

\bottomrule
\end{tabular}
}
\vspace{-2mm}
\caption{\textbf{Detailed hyperparameters of our \method.} LR denotes learning rate for training. Vision Embed. \& Head refers to the vision embedding and LM head of vision part.}
\vspace{-4mm}
\label{tab:hyperparams}
\end{table}
\begin{table}[t]
\center
\setlength{\tabcolsep}{3.6pt}
\Large
\resizebox{0.48\textwidth}{!}{
\begin{tabular}{lcc|cccc}
\toprule
 & & & \multicolumn{3}{c}{\it Emu Edit}  \\
\multirow{-2}[0]{*}{Method} & \multirow{-2}[0]{*}{Type} & \multirow{-2}[0]{*}{Tasks} & DINO & CLIP-I & CLIP-T \\
\midrule
InstructPix2Pix~\cite{brooks2023instructpix2pix} & Diffusion & Edit only & 0.762 & 0.834 & 0.219 \\
MagicBrush~\cite{zhang2024magicbrush} & Diffusion & Edit only & 0.776 & 0.838 & 0.222 \\
OmniGen~\cite{xiao2024omnigen} & Diffusion & Edit only & \underline{0.804} & 0.836 & 0.233 \\
Emu Edit~\cite{emuedit} & Diffusion & Edit only & \textbf{0.819} & 0.859 & 0.231\\
\midrule
PUMA~\cite{fang2024puma} & AR & Edit only & 0.785 & 0.846 & \textbf{0.270} \\
\rowcolor[gray]{0.9}ILLUME (Ours) & AR & Und, Gen, Edit & 0.791 & \textbf{0.879} & \underline{0.260} \\
\bottomrule
\end{tabular}
}
\vspace{-2mm}
\caption{\textbf{Quantitative results on image editing benchmarks. } The performance with top-1 and top-2 value are denoted in bold and underline respectively.
}
\vspace{-4mm}
\label{tab:editing_results}
\end{table}

\subsection{Evaluation Setup}

\paragraph{Multimodal Understanding.} To evaluate the multimodal understanding capabilities, we conduct evaluation on two types of widely-used benchmarks: (1) \textit{General}, including POPE~\cite{POPE}, MMBench~\cite{mmbench}, SEED~\cite{seed}, MME-P~\cite{mme}, MM-Vet~\cite{mmvet}, MMMU~\cite{mmmu} and AI2D~\cite{ai2d}; (2) \textit{Document-oriented}, including VQA-text~\cite{vqatext}, ChartQA~\cite{chartqa}, DocVQA~\cite{docvqa}, InfoVQA~\cite{infovqa} and OCRBench~\cite{ocrbench}. 

\paragraph{Multimodal Image Generation.} To evaluate the multimodal visual generation capability of ILLUME, we use the MJHQ-30K~\cite{mjhq}, GenAI-bench~\cite{genaibench} and GenEval~\cite{geneval} benchmarks. For MJHQ-30K, we adopt the Fréchet Inception Distance (FID~\cite{fid}) metric on 30K generated images compared to 30K high-quality images, measuring the generation quality and diversity. GenAI-bench and GenEval are challenging text-to-image generation benchmarks designed to reflect the comprehensive generative abilities. 

\vspace{-2mm}
\paragraph{Multimodal Image Editing.} To assess the multimodal image editing capability of our method, we evaluate it on the Emu Edit~\cite{emuedit} benchmark and report the CLIP-I, CLIP-T, and DINO~\cite{dino} scores. The CLIP-I and DINO scores measure the model’s ability to preserve elements from the source image, while the CLIP-T score measures the consistency between the output image and the target caption.

\subsection{Comparison with State-of-the-arts}
\paragraph{Multimodal Understanding.} We report the performances on various multimodal understanding benchmarks of \method~and previous state-of-the-art multimodal understanding-only models, including InstructBLIP~\cite{instructblip}, Qwen-VL-Chat~\cite{bai2023qwen}, LLaVA-1.5~\cite{liu2024improved}, ShareGPT4V~\cite{chen2023sharegpt4v}, LLaVA-NeXT~\cite{liu2024llava} and Emu3-Chat~\cite{wang2024emu3}, and unified models, including Unified-IO 2~\cite{lu2024unified}, Chameleon~\cite{team2024chameleon}, LWM~\cite{liu2024world}, Show-o~\cite{xie2024show}, VILA-U~\cite{wu2024vila}, and Janus~\cite{wu2024janus}, in Table~\ref{tab:understanding_results}. As we can see, ILLUME wins the first or second places on 10 out of 12 benchmarks. Specifically, ILLUME achieves $25\%$ and $14\%$ improvements on the MMMU and SEED benchmarks against the previous best unified multimodal model, Janus. Compared with the Emu3 model, ILLUME achieves comparable performance on document-oriented benchmarks and better performance on almost all general benchmarks, indicating the superiority of \method. 

\paragraph{Multimodal Image Generation.} We benchmark the multimodal image generation capability of ILLUME on MJHQ30K, GenAI-bench and GenEval benchmarks in Table \ref{tab:generation_results}. We compare ILLUME against previous state-of-the-art multimodal generation-only models, including SDv1.5~\cite{rombach2022high}, PixArt-$\alpha$~\cite{chen2023pixart}, SDXL~\cite{podell2023sdxl} and Emu3-Gen~\cite{wang2024emu3}, and unified models stated above. As we can see, ILLUME achieves 7.76 FID scores on the MJHQ30K benchmark, which is better than previous high-performance unified models such as Show-o and Janus, indicating better generation quality and diversity of ILLUME. We also achieve comparable results on the GenAI-bench benchmark against baseline methods and achieves the best overall accuracy (0.61) on the GenEval benchmark, surpassing previous generation-only and unified models, demonstrating superior comprehensive generation ability of ILLUME. 

\paragraph{Multimodal Image Editing.} We compare ILLUME with previous state-of-the-art multimodal image editing models including InstructPix2Pix~\cite{brooks2023instructpix2pix}, MagicBrush~\cite{zhang2024magicbrush}, OmniGen~\cite{xiao2024omnigen}, Emu Edit~\cite{emuedit} and PUMA~\cite{fang2024puma} in Table \ref{tab:editing_results}. 
As we can see, in comparison with those baseline models only supporting image editing task, ILLUME achieves competitive results even though it is a unified model, indicating the effectiveness of our framework.

\subsection{Ablation Studies}
\paragraph{Design Choice of Vision Tokenizer.}

\begin{figure}[t]
    \begin{center}
        \includegraphics[width=1.0\linewidth]{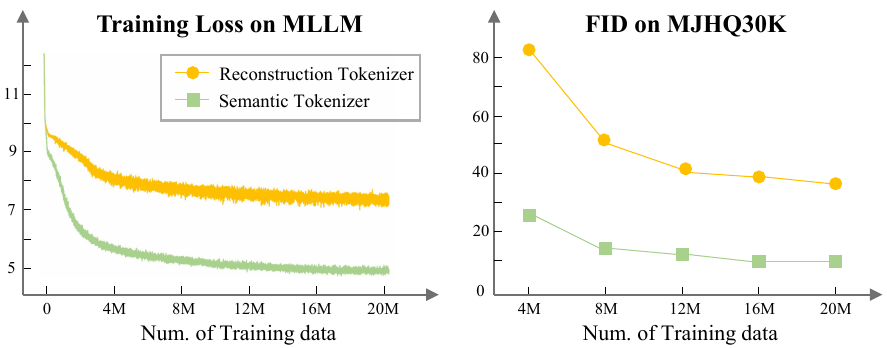}
    \end{center}
    \vspace{-4mm}
    \caption{\textbf{Comparison of different tokenizers for MLLM training.} We compare two types of tokenizers: 1) Reconstruction tokenizer: supervised by image reconstruction loss. 2) Semantic tokenizer: supervised by feature reconstruction loss. The results manifest that vision tokenizer with semantics significantly accelerates the convergence of MLLM pretraining. 
    }
    \label{fig-preliminary_tokenizer}
    \vspace{-4mm}
\end{figure}

To investigate whether semantic information is a pivotal factor in designing an effective vision tokenizer, we conduct a comparative analysis of vision tokenizers under two types of supervision:
1) \textit{Reconstruction tokenizer:} We use VQGAN as implemented in ~\cite{lee2022autoregressive}, which is supervised using image reconstruction loss. 2) \textit{Semantic tokenizer:} The quantization process is supervised with the objective of reconstructing semantic features extracted by UNIT~\cite{zhu2024unit}. We trained the MLLM with these two different tokenizers on 20M text-to-image generation dataset under the same setting. As depicted in Figure~\ref{fig-preliminary_tokenizer} (Left), the training-loss curves manifest that the vision tokenizer with semantics significantly hastens the MLLM training convergence.
For image reconstruction and detail compensation in our vision tokenizer, we employed a diffusion model. To ensure a fair comparison, we substituted the original decoder in VQGAN with a diffusion model to reconstruct images at $512\times512$ resolution. 
As shown in Figure~\ref{fig-preliminary_tokenizer} (Right), with only 20M generation data, the performance of the reconstruction tokenizer was unsatisfactory, whereas the semantic tokenizer achieved commendable performance. 
These findings confirm that semantic information is indeed a pivotal factor for a vision tokenizer suited to MLLMs.

\vspace{-2mm}
\paragraph{Effectiveness of Self-Enhancing Multimodal Alignment Scheme.}
In our study, we conducted an ablation analysis to validate the effectiveness of our approach. The baseline involves sampling a subset of 1.3M data points during Stage-3 training for efficiency, whereas our method also integrates assessment data generated by our scheme. 
As shown in Table~\ref{tab:ablation_analysis}, despite incorporating only 50K additional data, we observe improvements in performance across both understanding and generation benchmarks. This enhancement underscores that teaching the MLLM to self-assess not only enables the model to interpret images more accurately but also helps prevent potential errors in image generation. We hope this finding will inspire further exploration into the synergistic and generalization potentials between discriminative and generative capabilities.

\begin{table}[t]
\center
\setlength{\tabcolsep}{3.6pt}
\Large
\resizebox{0.48\textwidth}{!}{
\begin{tabular}{lccccccc}
\toprule
\multicolumn{8}{c}{\textbf{\it Understanding}} \\
\midrule
 &  POPE & MME-P & MMBench & SEED & GQA & MM-Vet & MMMU \\
baseline & 86.4 & 1358.6 & 61.7 & 65.0 & 60.0 & 27.4 & 31.2 \\
+ assessment & 86.1 & 1446.7 & 63.1 & 66.0 & 60.7 & 29.0 & 32.0  \\
\midrule
\multicolumn{8}{c}{\textbf{\it Generation}} \\
\midrule
 & Overall & Single Obj & Two Obj. & Counting & Colors & Position & Color Attri. \\
baseline & 0.56 & 0.98 & 0.8 & 0.35 & 0.69 & 0.34 & 0.22 \\
+ assessment & 0.59 & 0.99 & 0.84 & 0.43 & 0.72 & 0.33 & 0.24  \\
\bottomrule
\end{tabular}
}
\vspace{-2mm}
\caption{\textbf{Ablation of self-enhancing multimodal alignment.} }
\vspace{-4mm}
\label{tab:ablation_analysis}
\end{table}

\section{Conclusion}
In this paper, we introduce \method, a unified MLLM which is efficiently pretrained and further improved by a novel self-enhancing multimodal alignment scheme, exhibiting competitive or even superior performance compared to existing unified MLLMs across various multimodal benchmarks. 
Looking ahead, we plan to further develop \method ~in several key areas: 
1) We aim to extend its capabilities to accommodate more modalities,
such as video, audio and 3D data, for a broader applicability across various fields. 
2) We intend to design a more versatile vision tokenizer that can support both images and videos. Moreover, our findings in this study suggest that incorporating semantic information into traditionally well-designed vision tokenizers holds great potential for making them more suitable for MLLMs.
3) We plan to further explore our self-enhancing strategy by incorporating more recognized criteria, such as aesthetic quality, allowing for better data utilization and generation that more closely align with human preferences.
These future directions will significantly broaden the applicability and effectiveness of \method, paving the way for a unified, highly effective, and efficient any-task, any-modality MLLM.

\clearpage
{
    \small
    \bibliographystyle{ieeenat_fullname}
    \bibliography{arxiv}
}

\newpage
\appendix
\renewcommand{\thetable}{\Alph{table}}
\renewcommand{\thefigure}{\Alph{figure}}
\renewcommand{\thesection}{\Alph{section}}

\setcounter{figure}{0}
\setcounter{table}{0}

\section{Implementation Details of Vision Tokenizer}
We supervise the quantization process within a semantic feature space, which is promising to facilitate the image-text alignment in MLLM training. 
Given an image $x$, it is fed into UNIT encoder to extract semantic features $X=\{x_1, ..., x_N\}$. The semantic features then pass into a quantizer, which tokenizes $X$ to a sequence of discrete tokens $V=\{v_1, ..., v_N\}$ by looking up a learnable codebook $\mathcal{C}=\{c_1,... c_K\}$, where $K$ is the codebook size. The discrete token $v_i$ is calculated by assigning $x_i$ to its closest neighbourhood code in $\mathcal{C}$ according to the L2 norm:
\begin{equation}
    v_i = \mathop{\arg\min}\limits_{j} ||x_i - c_j||, v_i \in [0, K-1]
\end{equation}
Based on the discrete tokens, we can obtain its quantized embeddings, which is then fed into a decoder to obtain reconstructed semantic features $X^{rec}=\{x_1^{rec}, ..., x_N^{rec}\}$. The quantization process is supervised by the feature reconstruction loss using \textit{cosine loss} and \textit{smoothl1 loss}:
\begin{equation}
    \mathcal{L} = \sum_{i=1}^{N} (smoothl1(x_i, x_i^{rec}) + (1 - cosine(x_i, x_i^{rec})))
\end{equation}
During training, the vision encoder is kept frozen and only the parameters of quantizer and decoder are updated. It is trained for 80K steps on 80M images with the batch size of 2048, epochs of 2 and learning rate of 5e-5.

To further recover the original pixel space, the reconstructed semantic features are set as conditions and injected to each block of a conditional diffusion model through cross-attention layers. The conditional U-Net is initialized from  SDXL and finetuned 80K steps with the batch size of 128 and learning rate of 2e-5. Only the attention layer of U-Net is updated for efficient training. Note that the whole tokenizer training only requires pure image data without corresponding text descriptions.

\section{More Results of ILLUME}

\begin{figure*}[t]
    \begin{center}
        \includegraphics[width=0.9\linewidth]{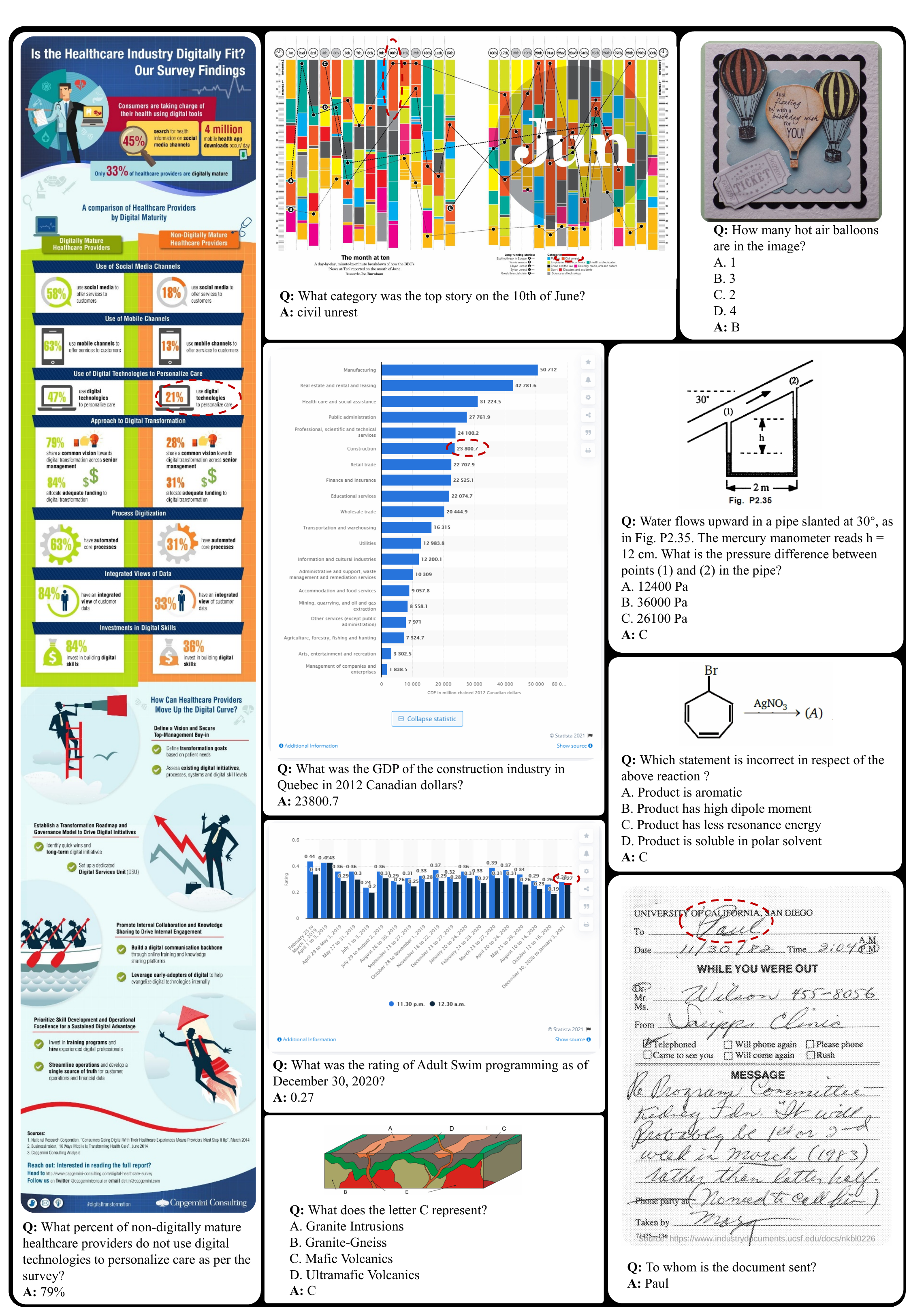}
    \end{center}
    \vspace{-5mm}
    \caption{More qualitative results on understanding tasks. Regions that related to the QAs are marked with red ellipses.
    }
    \label{fig-more_vis_und}
    \vspace{-4mm}
\end{figure*}

\begin{figure*}[t]
    \begin{center}
        \includegraphics[width=1.0\linewidth]{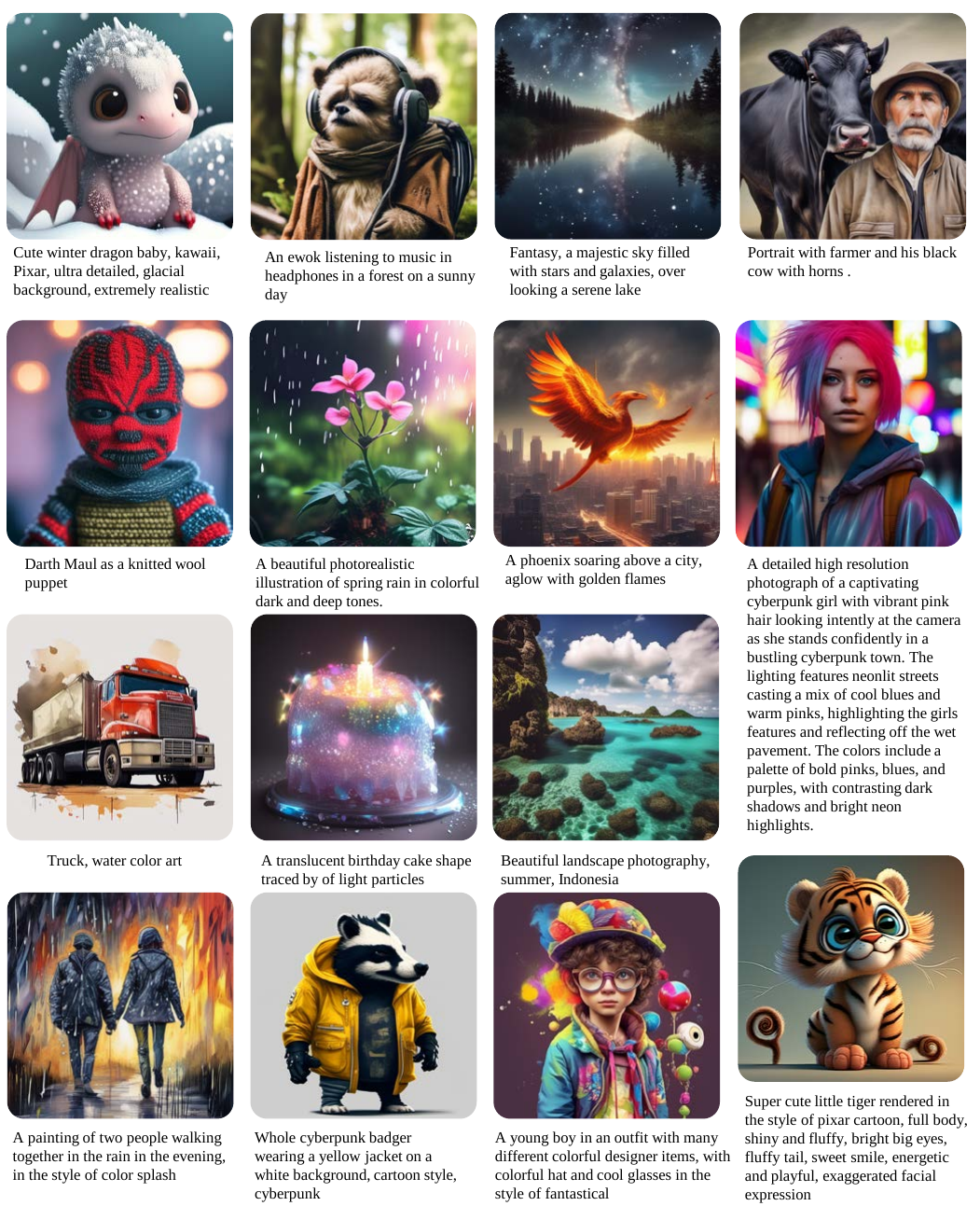}
    \end{center}
    \vspace{-5mm}
    \caption{More qualitative results on text-to-image generation tasks.
    }
    \label{fig-more_vis_gen}
    \vspace{-4mm}
\end{figure*}

\begin{figure*}[t]
    \begin{center}
        \includegraphics[width=0.9\linewidth]{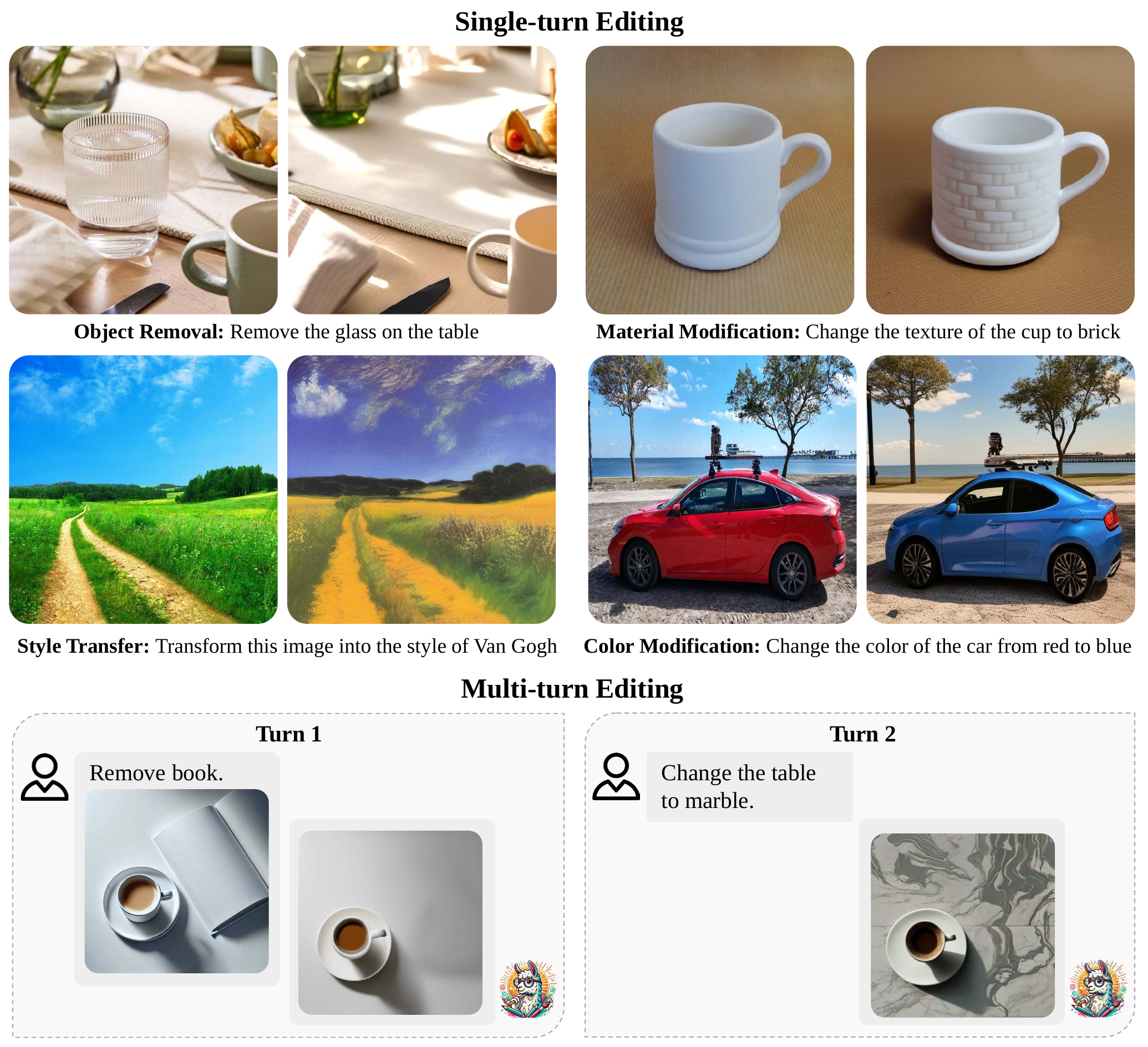}
    \end{center}
    \vspace{-5mm}
    \caption{More qualitative results on mixed-modal generation tasks.
    }
    \label{fig-more_vis_edit}
    \vspace{-4mm}
\end{figure*}

\paragraph{More qualitative results.} 
Figure~\ref{fig-more_vis_und} showcases additional qualitative results for comprehension tasks, demonstrating that our ILLUME model can adeptly handle various comprehension tasks and images with significant differences in aspect ratio. 
Figures~\ref{fig-more_vis_gen} and Figures~\ref{fig-more_vis_edit} provide further visualizations in text-to-image generation and mixed-modal generation tasks, respectively. 
In the future, we plan to enhance MLLMs to produce higher resolution images and to support a wider range of mixed-modal generation tasks.

\paragraph{Detailed performance results on GenAI-bench.}
We details per-category performance on GenAI-bench in Table~\ref{tab:detail_results_genaibench}, where our ILLUME achieves competitive results with current autoregressive-based unified MLLMs.

\begin{figure}[t]
    \begin{center}
        \includegraphics[width=1.0\linewidth]{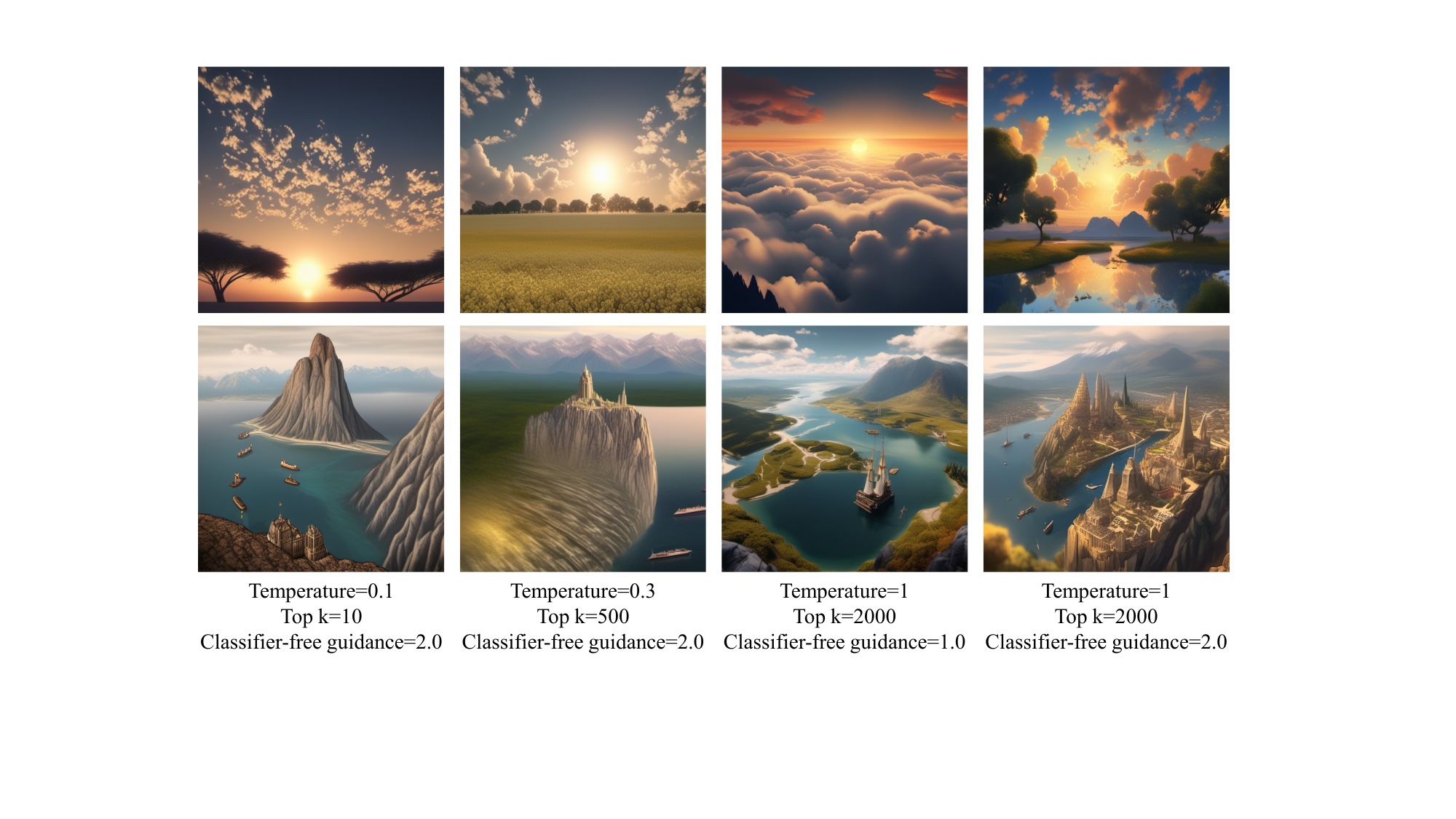}
    \end{center}
    \vspace{-6mm}
    \caption{Comparison of different hyper-parameters in inference.}
    \label{fig-inference}
    \vspace{-6mm}
\end{figure}

\paragraph{Inference hyper-parameters.}
Figure~\ref{fig-inference} presents a comparison of different inference decoding hyperparameters for text-to-image generation. It can be observed that increasing temperature, top-k, and guidance scale all lead to improved visual details.

\begin{figure*}[b]
    \begin{center}
        \includegraphics[width=0.9\linewidth]{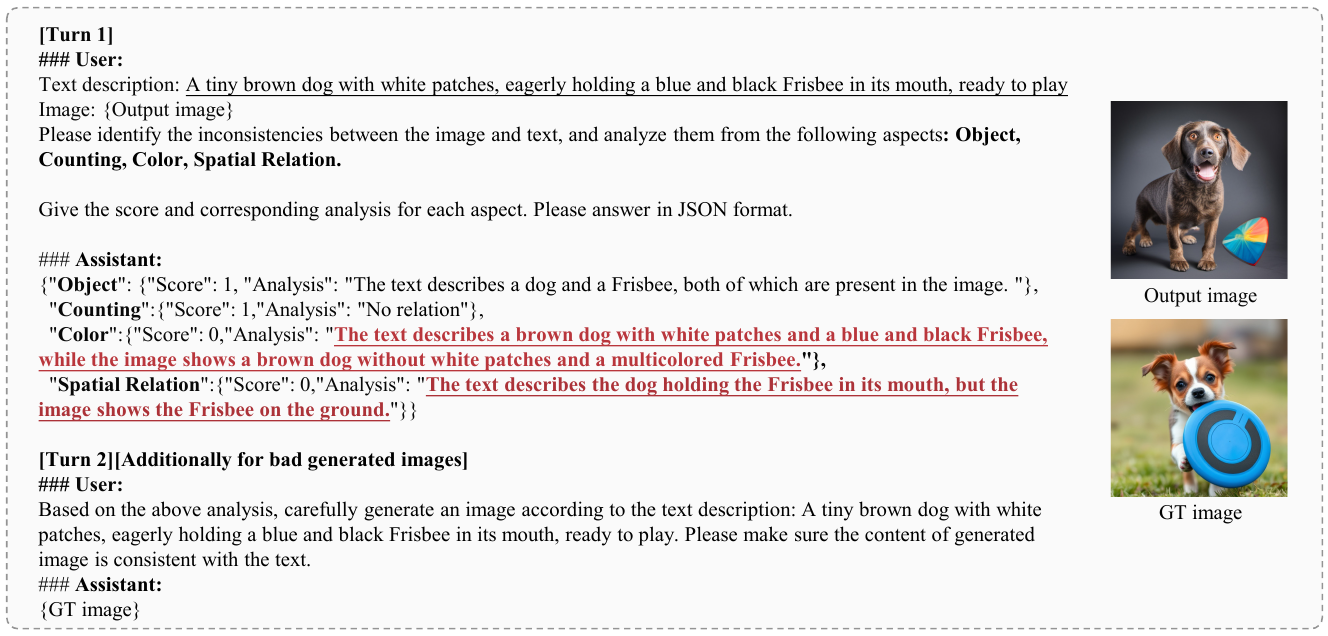}
    \end{center}
    \vspace{-5mm}
    \caption{Data example of assessment data for self-enhancing multimodal alignment.
    }
    \label{fig-assessment_data}
    \vspace{-4mm}
\end{figure*}

\paragraph{Data examples of assessment data.}
Figure~\ref{fig-assessment_data} illustrates an example of assessment data for self-enhancing multimodal alignment scheme. This example showcases how the data identifies specific reasons for inconsistencies between self-generated images and text descriptions, which aids the model in interpreting images more accurately and helps prevent mistakes during image generation.

\begin{table*}[hb]
\center
\setlength{\tabcolsep}{3.6pt}
\Large
\resizebox{\textwidth}{!}{
\begin{tabular}{lcc|cccccc|cccccc}
\toprule
 &   &  & \multicolumn{6}{c|}{\it Basic} & \multicolumn{6}{c}{\it Advanced} \\
\multirow{-2}[0]{*}{Method} & \multirow{-2}[0]{*}{Params.} & \multirow{-2}[0]{*}{Type} & Attribute & Scene & Spatial & Action & Part & Overall & Count & Differ & Compare & Negate & Universal & Overall \\
\midrule
SDXL & 2.6B & Diffusion & 0.84 & 0.84  & 0.82 & 0.83 & 0.89 & 0.83 & 0.71 & 0.73  & 0.69 & 0.50 & 0.66 & 0.63 \\
LWM & 7B & Autoregressive & 0.63 & 0.62 & 0.65 & 0.63 & 0.70 & 0.63 & 0.59 & 0.58  & 0.54  & 0.49 & 0.52 & 0.53 \\
Show-o &  1.5B & Autoregressive & 0.72 & 0.72 & 0.70 & 0.70 & 0.75 & 0.70 & 0.70 & 0.62  & 0.71  & 0.51 & 0.65 & 0.60 \\
VILA-U(256)&  7B & Autoregressive & 0.78 & 0.78 & 0.77 & 0.78 & 0.79 & 0.76 & 0.70 & 0.71  & 0.74  & 0.53 & 0.66 & 0.64 \\
VILA-U(384)& 7B & Autoregressive & 0.75 & 0.76 & 0.75 & 0.73 & 0.75 & 0.73 & 0.68 & 0.67  & 0.71 & 0.51 & 0.64 & 0.61 \\
\midrule
ILLUME (Ours) & 7B & Autoregressive & 0.75 & 0.79 & 0.75 & 0.77 & 0.73 & 0.75 & 0.66 & 0.68 & 0.67 & 0.49 & 0.63 & 0.60 \\
\bottomrule
\end{tabular}
}
\vspace{-2mm}
\caption{Detailed quantitative results on GenAI-bench.
}
\vspace{-4mm}
\label{tab:detail_results_genaibench}
\end{table*}

\end{document}